\documentclass[acmsmall]{acmart}
\AtBeginDocument{%
  }

\setcopyright{acmlicensed}
\copyrightyear{2025}
\acmYear{2025}
\acmDOI{XXXXXXX.XXXXXXX}

\acmJournal{CSUR}
\acmVolume{37}
\acmNumber{4}
\acmArticle{111}
\acmMonth{6}

\usepackage{multirow}

\begin{document}

\title{When Large Language Models Meet Law: Dual-Lens Taxonomy, Technical Advances, and Ethical Governance}

\author{Peizhang Shao}
\authornote{Both authors contributed equally to this research.}
%\email{peizhang.shao@law.ox.ac.uk}
\affiliation{%
  \institution{School of Law, China University of Political Science and Law}
  \city{Beijing}
  \country{China}
}
\affiliation{%
  \institution{Zhejiang University of Finance and Economics Dongfang College}
  \city{Haining}
  \country{China}
}
\author{Linrui Xu}
\authornotemark[1]
%\email{webmaster@marysville-ohio.com}
\affiliation{
\institution{School of Information Management for Law, China University of Political Science and Law}  \city{Beijing}  \country{China} } \affiliation{
\institution{Department of Artificial Intelligence, Chung-Ang University}  \city{Seoul}  \country{Republic of Korea} }

\author{Jinxi Wang}
\affiliation{%
  \institution{School of Law, China University of Political Science and Law}
  \city{Beijing}
  \country{China}
}
%\email{jinxiw@cupl.edu.cn}

\author{Wei Zhou}
\authornote{Corresponding Authors: Wei Zhou and Xingyu Wu}
\email{cu0085859@cupl.edu.cn}
\affiliation{%
  \institution{School of Information Management for Law, China University of Political Science and Law}
  \city{Beijing}
  \country{China}
}
\affiliation{%
  \institution{School of Law, China University of Political Science and Law}
  \city{Beijing}
  \country{China}
}

\author{Xingyu Wu}
\authornotemark[2]
\email{xingy.wu@polyu.edu.hk}
\affiliation{%
 \institution{Department of Data Science and Artificial Intelligence, The Hong Kong Polytechnic University}
 \city{Hong Kong SAR}
 \country{China}}

\renewcommand{\shortauthors}{Shao et al.}

\begin{abstract}
This paper establishes the first comprehensive review of Large Language Models (LLMs) applied within the legal domain. It pioneers an innovative dual lens taxonomy that integrates legal reasoning frameworks and professional ontologies to systematically unify historical research and contemporary breakthroughs. Transformer-based LLMs, which exhibit emergent capabilities such as contextual reasoning and generative argumentation, surmount traditional limitations by dynamically capturing legal semantics and unifying evidence reasoning. Significant progress is documented in task generalization, reasoning formalization, workflow integration, and addressing core challenges in text processing, knowledge integration, and evaluation rigor via technical innovations like sparse attention mechanisms and mixture-of-experts architectures. However, widespread adoption of LLM introduces critical challenges: hallucination, explainability deficits, jurisdictional adaptation difficulties, and ethical asymmetry. This review proposes a novel taxonomy that maps legal roles to NLP subtasks and computationally implements the Toulmin argumentation framework, thus systematizing advances in reasoning, retrieval, prediction, and dispute resolution. It identifies key frontiers including low-resource systems, multimodal evidence integration, and dynamic rebuttal handling. Ultimately, this work provides both a technical roadmap for researchers and a conceptual framework for practitioners navigating the algorithmic future, laying a robust foundation for the next era of legal artificial intelligence.
\end{abstract}




\received{20 February 2025}
\received[revised]{12 March 2025}
\received[accepted]{5 June 2025}

\maketitle

\section{Introduction}

The field of Artificial Intelligence and Law has evolved through three transformative decades since the inaugural issue of its flagship conference ICAIL in 1987 \cite{araszkiewicz2022thirty}. Early research focused on symbolic approaches including legal ontologies for knowledge sharing \cite{gruber1991role}, rule-based reasoning systems, and argumentation frameworks. These established foundational paradigms but faced persistent challenges: ontological engineering struggled with semantic interoperability as conceptual mismatches hindered cross-system reuse \cite{visser1998comparison}, evidence reasoning remained fragmented between narrative and probabilistic methods without unified computational frameworks \cite{bex2003towards}, and practical applications rarely moved beyond laboratory prototypes due to knowledge engineering bottlenecks \cite{lauritsen1992technology}. This historical context illuminates why the evolution of legal artificial intelligence has transitioned from these symbolic expert systems \cite{susskind1996future} to contemporary data-driven neural models \cite{ashley2017artificial}, marking a paradigm shift from logic-based formalisms to statistical learning approaches capable of handling law's inherent complexities. In summary, Early AI systems grappled with two fundamental constraints rooted in law's unique characteristics. First, the semantic richness of legal language demands context-aware interpretation beyond syntactic patterns—a challenge acutely felt in ontology engineering where attempts to formalize concepts like "reasonable doubt" revealed irreconcilable jurisdictional variations \cite{peters2007structuring}. Second, the procedural rigidity of task-specific architectures proved ill-equipped for multi-step reasoning across evidentiary, normative, and factual dimensions \cite{palau2009argumentation, lippi2016argumentation}, as seen in isolated evidence models that combined arguments, stories, and Bayesian networks without theoretical integration \cite{vlek2016method}. While transitional technologies like domain-adapted embeddings (Law2Vec; \cite{chalkidis2017deep}) and BiLSTM-based judgment predictors \cite{chalkidis2019neural} offered incremental progress, their narrow task focus and limited generalization capacity fundamentally hindered end-to-end legal workflow integration. These limitations were particularly evident in real-world deployments where legal practitioners required systems capable of fluidly transitioning between statutory interpretation, precedent analysis, and evidentiary assessment—a holistic capability that remained elusive until the advent of large language models.

The emergence of transformer-based architectures has catalyzed a transformative shift in legal AI. Unlike predecessors, LLMs exhibit emergent capabilities—contextual reasoning, few-shot adaptation, and generative argumentation—that transcend traditional NLP pipelines \cite{brown2020language}. These capabilities directly address long-standing gaps identified during AI \& Law's first three decades: they overcome ontological fragmentation through contextual embeddings that dynamically capture legal semantics \cite{leone2020taking}, unify evidence reasoning via coherent narrative generation that integrates probabilistic and argumentative elements \cite{bex2013legal}, and enable practical deployment at scale by circumventing manual knowledge engineering bottlenecks \cite{oskamp2002ai}. This technological leap redefines the AI-Law nexus fundamentally. Where pre-LLM research optimized fragmented tasks—such as charge prediction via BiLSTMs \cite{zhong2020does} or statute retrieval using BM25 \cite{rabelo2019combining}—contemporary frameworks (GPT-4 \cite{achiam2023gpt}, LLaMA-Law \cite{cui2023chatlaw}) now enable unified architectures capable of handling end-to-end litigation and non-litigation workflows.

Crucially, the legal domain demands LLMs due to their unprecedented alignment with three core requirements that have challenged legal informatics since its inception. For complex text processing, traditional methods like rule-based summarization and SVM classification failed to capture jurisprudential nuance—a limitation starkly evident in early lexical ontologies \cite{peters2007structuring}. LLMs overcome this through abstractive summarization techniques like the ETA method \cite{jain2024domain} that preserve legal semantics, though they introduce hallucinations mitigated via knowledge-graph grounding architectures such as ChatLaw \cite{cui2024chatlaw}. Regarding reasoning and argumentation, symbolic systems implementing Toulmin frameworks \cite{toulmin2003uses} lacked scalability—a gap documented in Walton's argumentation schemes \cite{walton2010similarity}. Modern LLMs enable generative warrant reasoning through specialized architectures like Lawformer \cite{xiao2021lawformer} while leveraging retrieval-augmented generation (RAG) for evidentiary backing in COLIEE benchmarks \cite{goebel2024overview}, though qualifier calibration in sentencing prediction remains challenging \cite{lyu2022improving}. Finally, for procedural augmentation, where pre-LLM tools operated in isolation (e.g., mediation outcome predictors \cite{boella2011using}), contemporary systems facilitate human-AI collaboration in multi-agent negotiation scenarios \cite{yue2023disc}, albeit raising ethical concerns about adversarial asymmetry originally identified in legal evidence studies \cite{bex2003towards}.

This paper systematically summarizes three synergistic approaches essential for the effective deployment of LLMs in the demanding domain of legal practice. These approaches address fundamental challenges that have historically hindered legal AI systems and collectively enable transformative capabilities. First, context scalability—absolutely critical for processing the thousand-page case files commonplace in litigation—is achieved through advanced sparse attention mechanisms exemplified by architectures like Longformer \cite{beltagy2020longformer}. This technical innovation directly overcomes the evidence integration bottlenecks identified in earlier hybrid Bayesian models \cite{vlek2014building}, allowing LLMs to efficiently analyze vast evidentiary records that were previously computationally intractable. The ability to handle extensive context is foundational for tasks like comprehensive case summarization and identifying subtle precedents buried within lengthy documents, moving beyond the fragmented capabilities of earlier systems. Second, knowledge integration—crucial for grounding LLM outputs in authoritative legal principles and drastically reducing hallucination—is implemented via sophisticated hybrid architectures. Techniques such as mixture-of-experts (MoE) systems integrated with structured legal knowledge graphs, as demonstrated in models like ChatLaw \cite{cui2024chatlaw}, provide this essential grounding. This approach represents a significant evolution, extending the foundational ontology engineering principles established by pioneers like Visser and Bench-Capon \cite{visser1998comparison} into the neural network era. By explicitly incorporating domain-specific knowledge structures, these systems enhance the reliability of outputs for critical tasks like legal reasoning and argument generation, ensuring fidelity to established legal doctrine where earlier AI often faltered. Third, evaluation rigor—an aspect long neglected in the deployment of practical legal AI systems \cite{lauritsen1995technology}—is now being systematically addressed through the adoption of specialized, domain-relevant benchmarks. Frameworks like LawBench \cite{fei2023lawbench} and LexGLUE \cite{chalkidis2021lexglue} establish standardized performance metrics tailored to legal tasks, moving beyond generic NLP evaluations. This rigorous assessment paradigm is vital for quantifying progress in complex legal applications such as judgment prediction, legal question answering, and precedent retrieval, providing the necessary evidence base for trust and adoption in professional settings. These three synergistic approaches—context scalability, knowledge integration, and evaluation rigor—collectively underpin three paradigm-shifting advancements that bridge persistent historical gaps in legal AI. 

Despite these transformative advances, the LLM revolution introduces novel challenges that demand scholarly attention and technical innovation. Hallucination in legal claims manifests as spurious citations or normative fabrications, with cross-jurisdictional question answering systems exhibiting error rates as high as $58\%$ \cite{dahl2024large}—a phenomenon echoing computational trust issues in expert testimony identified by Walton \cite{walton2003there}. Mitigation strategies like retrieval augmentation \cite{nguyen2024pushing} show promise but require further refinement. Explainability deficits create significant accountability gaps as black-box reasoning obscures decision pathways, perpetuating adoption barriers first noted in the 1990s \cite{hokkanen2002knowledge} and partially addressed through syllogistic prompting techniques \cite{deng2023syllogistic}. Jurisdictional adaptation remains problematic, with performance degradation in low-resource legal systems \cite{maree2024transforming} necessitating specialized few-shot tuning approaches that extend ontological localization challenges documented by Leone et al. \cite{leone2020taking}. Perhaps most critically, ethical asymmetry emerges when disparities in LLM access exacerbate existing power imbalances between legal actors \cite{weidinger2021ethical}, requiring fairness-aware architectural interventions that respond to Walton's foundational concerns about argumentative justice \cite{walton2014baseballs}. These challenges collectively represent the "next frontier" in legal AI, demanding interdisciplinary solutions that integrate technical innovation with jurisprudential principles.

Previous reviews in the field of law and artificial intelligence have two main significant issues: On the one hand, there is the issue of comprehensiveness. Some works only focus on subtasks in the NLP domain \cite{ariai2024natural, yang2024large, anh2023impact}, subtasks in the judicial prediction domain \cite{cui2023survey, feng2022legal}, or subtasks in the IR domain \cite{nguyen2025retrieve, martinez2023survey}, which results in the review content often being insufficiently comprehensive. On the other hand, there is the issue of depth. They do not consider the legal problem itself, fail to take into account the ontological development process of legal artificial intelligence (or the ontological issues) \cite{lai2024large}, lack a technical review and large model application framework based on legal reasoning itself \cite{siino2025exploring}, or the articles themselves are relatively brief \cite{sun2023short, qin2024exploring}. Moreover, there is no review or discussion of the rules for legal professionals to use large models and their practical application areas.

This work presents the first comprehensive review of LLMs in legal contexts through a novel dual-lens taxonomy that integrates historical insights with contemporary breakthroughs. Our contributions advance the field in three significant dimensions:
\begin{itemize}
    \item \textbf{Dual Lens Taxonomy Innovation}: Inspired by Calegari et al.'s work about legal reasoning \cite{calegari2019defeasible}, Section 3 establishes the first legal reasoning ontology framework aligning Toulmin's argumentation components (Data, Warrant, Backing, Claim) with LLM workflows, implementing Bex's evidence theory \cite{bex2011arguments} at scale, while incorporating contemporary LLM breakthroughs with the evidence research of the past \cite{verheij2017proof} ; Section 4 mapping practitioner roles (lawyers, judges, litigants) to NLP tasks, extending user-centered ontology studies initiated by Francesconi and Araszkiewicz \cite{francesconi2014description}.
    
    \item \textbf{Scenario-Adaptive Deployment Framework}: Section 4 pioneers a role-centric framework for embedding LLMs across litigation (Section 4.1) and non-litigation (Section 4.2) workflows, addressing Lauritsen's call for \textit{"smarter tools"} \cite{lauritsen1992technology}.
    
    \item \textbf{Ethical \& Professional Mapping}: Section 5 presents the first systematic analysis of ethical challenges across practitioner roles (lawyers, judges, litigants), addressing both technological ethics (bias mitigation, hallucination controls) and legal professional responsibilities \cite{francesconi2014description}, extending user-centered ontology for LLM deployment \cite{garcia2024ontology}.
\end{itemize}

\begin{figure}
  \centering
  \includegraphics[width=\textwidth]{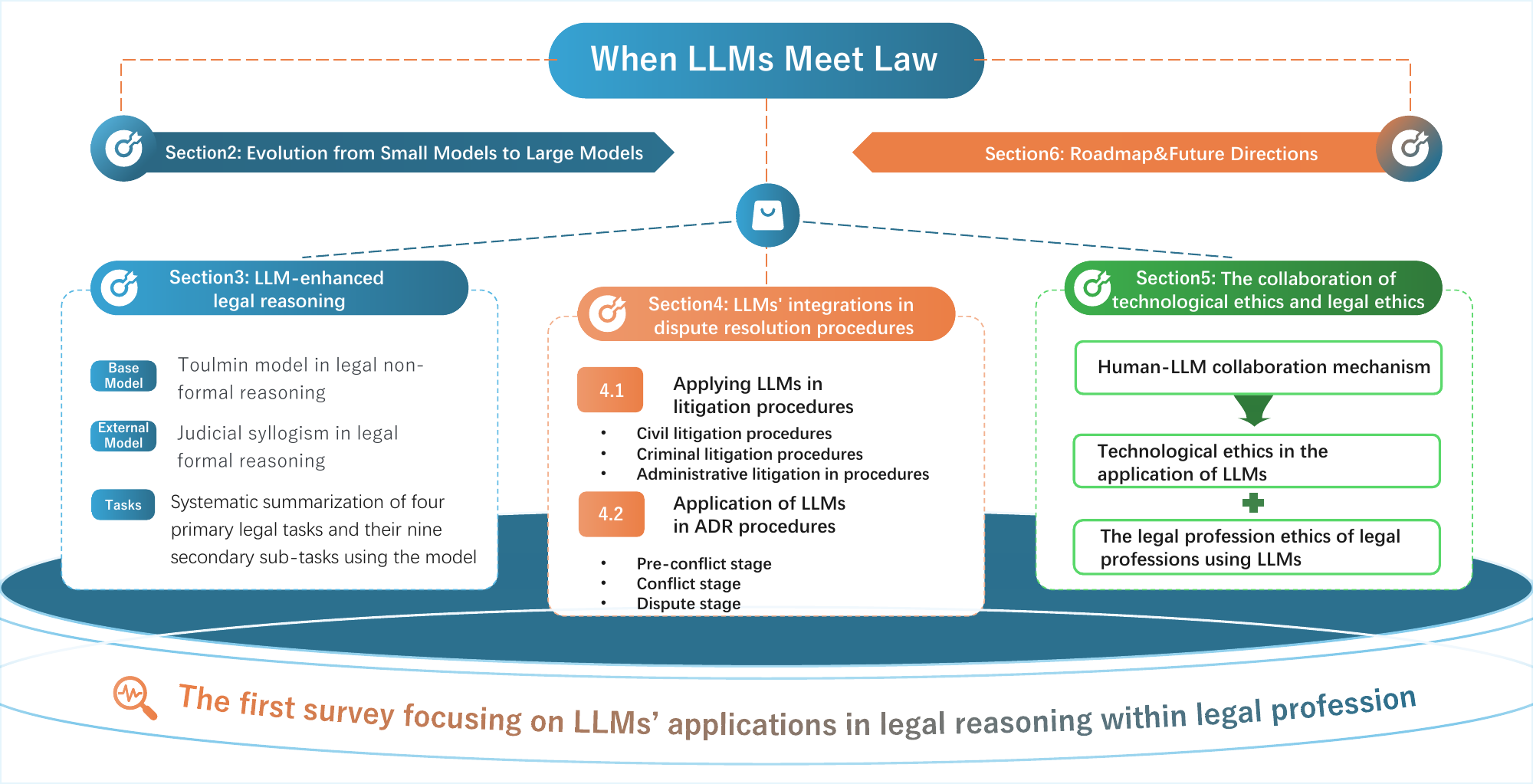}
  \caption{A general framework for integrated research of LLMs and Law}\label{Fig_Intro}
\end{figure}

As shown in Figure \ref{Fig_Intro}, the remainder of this paper is organized as follows. We first trace the technical evolution from small models to LLMs in Section 2, examining how neural architectures overcome knowledge representation barriers identified in early ontological studies. Section 3 analyzes LLM-enhanced legal reasoning through the Toulmin framework, extending legal LLMs tasks to the the ontology of legal reasoning. Next, Section 4 examines human-AI collaboration in litigation and alternative dispute resolution. Finally, Section 5 discusses ethical co-regulation frameworks and future directions in the era of LLM, incorporating lessons from three decades of applied research in legal technology. We have created a GitHub repository to index the relevant papers: \url{https://github.com/Kilimajaro/LLMs_Meet_Law}

\section{Evolution from Small Models to Large Models}

Section 2 traces the evolution of legal AI, contextualizing its progression from task-specific small models to versatile LLMs within broader NLP advancements. Section 2.1 outlines the technical trajectory of NLP, beginning with static embeddings (e.g., Word2Vec) and advancing through transformer architectures (e.g., BERT, GPT-3) that enable scalable, context-aware reasoning. Section 2.2 then examines how legal AI mirrored this shift, transitioning from early systems such as domain-adapted embeddings (Law2Vec) and BiLSTMs for clause segmentation to LLM-driven frameworks (Legal-BERT, LexGLUE) and generative models (e.g., GPT-3, ChatLaw) fine-tuned for legal tasks. %Together, these sections highlight how computational scalability and domain-specific adaptations—such as hybrid retrieval-augmented generation (RAG)—have bridged general NLP capabilities with the unique demands of legal language, fostering systems that balance versatility with legal precision.

\subsection{Evolution of NLP Models}

The evolution of NLP models has been characterized by a paradigm shift from small-scale, task-specific architectures to large-scale, general-purpose frameworks \cite{wu2024evolutionary}. Early advancements in distributed semantic representation laid the groundwork for modern language models. The introduction of word2vec by \cite{mikolov2013distributed} marked a pivotal transition, enabling efficient learning of word embeddings through skip-gram and continuous bag-of-words (CBOW) architectures. These embeddings capture syntactic and semantic relationships, forming the basis for downstream NLP tasks. However, the limitations of static embeddings in handling polysemy and contextual nuances have spurred the development of dynamic, context-aware architectures. Subsequent innovations such as ERNIE \cite{zhang2019ernie} extended this paradigm by incorporating structured knowledge graphs through entity alignment and novel pretraining objectives, demonstrating significant improvements in knowledge-intensive tasks such as entity typing and relation classification through bidirectional fusion of textual and knowledge representations.

The architectural foundations of modern NLP were established by the transformer model \cite{vaswani2017attention}, which introduced self-attention mechanisms to replace sequential processing, enabling parallelized training and superior long-range dependency modeling. Two distinct pretraining paradigms emerged: bidirectional encoding and autoregressive generation. The BERT model \cite{devlin2019bert} pioneered masked language modeling (MLM) through bidirectional transformer encoders, generating context-aware representations that revolutionized tasks such as question answering. Concurrently, the GPT series \cite{radford2018improving, radford2019language} established the autoregressive paradigm, with GPT-1 using left-to-right language modeling on BooksCorpus and GPT-2 scaling to 1.5B parameters on WebText data, demonstrating zero-shot task transfer capabilities. This dichotomy was bridged by hybrid architectures such as BART \cite{lewis2019bart}, which combines bidirectional encoders with autoregressive decoders through denoising objectives (e.g., text infilling and sentence permutation), achieving state-of-the-art performance in both comprehension and generation tasks. Moreover, efficiency-driven innovations emerged: ELECTRA \cite{clark2020electra} replaced MLM with replaced token detection, achieving 4× faster convergence by training discriminators to distinguish real tokens from generator-produced substitutes, whereas ALBERT \cite{lan2019albert} reduced parameter counts through factorized embedding layers and cross-layer parameter sharing without compromising performance.

The scaling hypothesis reached its zenith with GPT-3 \cite{brown2020language}, whose 175B parameters and few-shot learning capabilities redefined model generalization, whereas the T5 framework \cite{raffel2020exploring} unified all NLP tasks as text-to-text transformations, pretraining on a 34B token C4 corpus with span corruption objectives. Notably, ERNIE 3.0 \cite{sun2021ernie} integrated these directions through a hybrid architecture combining autoregressive flow and autoencoding paradigms, further enhanced by knowledge graph embeddings across 96 relation types, achieving $85.9\%$ accuracy on Chinese open-domain QA—a $7.2\%$ absolute gain over pure text-based models.

\subsection{Legal Domain Adaptation Trajectory}

The evolution of artificial intelligence in legal practice reflects broader technological trends in natural language processing, progressing from task-specific small models to versatile LLMs. Early efforts focused on addressing the unique challenges of legal texts—such as domain-specific terminology, syntactic complexity, and structured reasoning requirements—through tailored architectures and rule-based systems. These foundational approaches emphasized feature engineering and domain adaptation, as seen in the development of legal word embeddings and specialized recurrent neural networks. The emergence of transformer-based architectures and pretraining paradigms catalyzed a shift toward scalable, generalizable models capable of handling diverse legal tasks.

This subsection reviews landmark literature on the methodological evolution of core legal NLP applications, as illustrated in Fig. 2. By contrasting traditional techniques with contemporary LLM-driven approaches, it highlights pivotal studies that marked paradigm shifts in the field.

\begin{figure}
  \centering
  \includegraphics[width=\textwidth]{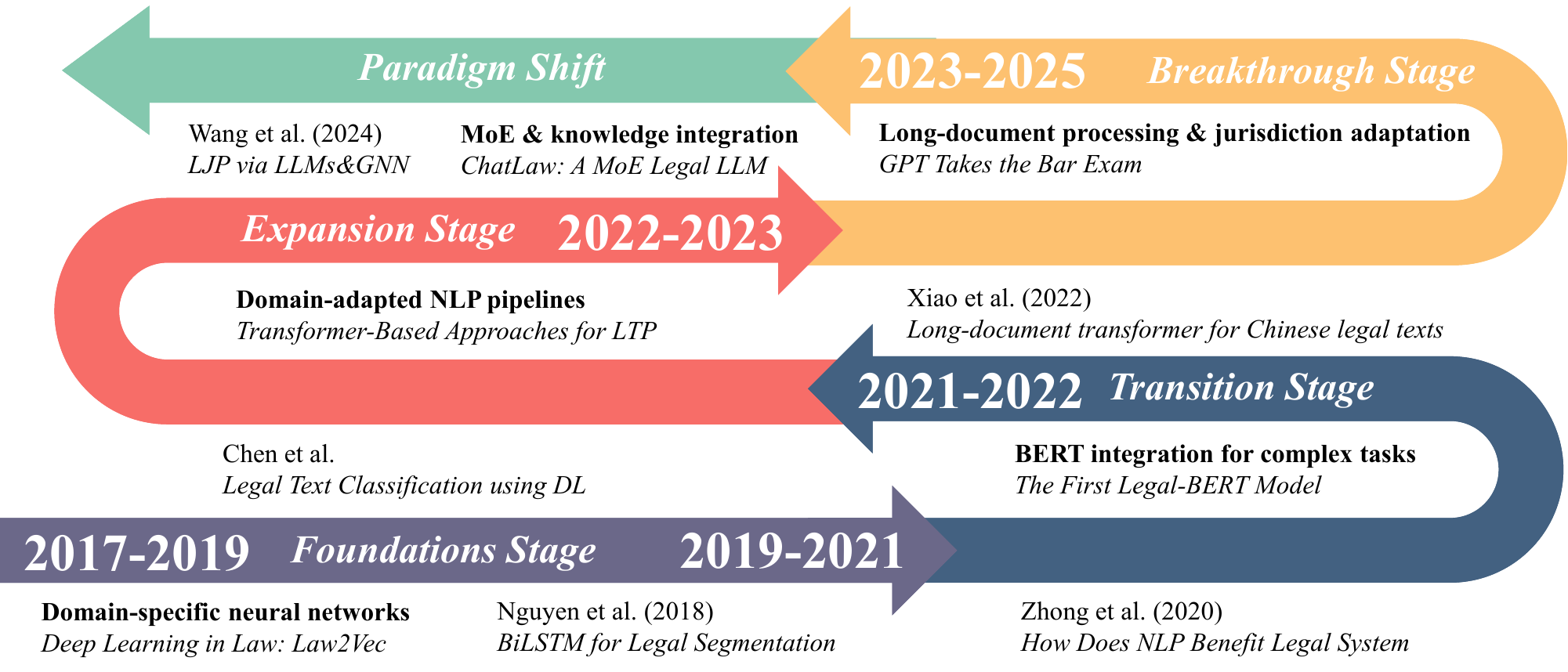}
  \caption{Evolution of Legal NLP Models: From Task-Specific Small Models to Large Models Era}\label{Fig_History}
\end{figure}

\textbf{Foundations Stage (2017-2019)} witnessed pioneering domain-specific architectures. Representative works include: Chalkidis et al.'s Law2Vec established legal word embeddings through neural networks, directly addressing lexical gaps in general-purpose models \cite{chalkidis2017deep}. Nguyen et al. refined BiLSTM , showing hybrid approaches where legacy architectures could integrate BERT-derived features for complex tasks like legal discourse segmentation \cite{nguyen2018recurrent}.

\textbf{Transition Stage (2019-2021)} marked the paradigm shift to pretraining. Chalkidis Legal-BERT became the cornerstone for transfer learning, repurposing transformer pre-training on case law corpora to capture jurisdictional semantics \cite{chalkidis2019neural}. And Zhong et al. systematized how domain-adapted NLP pipelines improve legal workflows \cite{zhong2020does}.

\textbf{Expansion Stage (2021-2022)} saw the transformer specialization scaling new complexities. Nguyen et al. developed a transformer-based approaches for the legal text processing task (LTP) in the COLIEE 2021 competition \cite{nguyen2022transformer}, while Chen et al. compared the random forests with deep learning in legal text classification tasks for the first time \cite{chen2022comparative}.

\textbf{Breakthrough Stage (2022-2023)} culminated in holistic LLM applications transcending task boundaries: Bommarito et al.'s Bar Exam GPT validated generative models' comprehensive legal reasoning \cite{bommarito2022gpt}; while Xiao et al.'s long-document transformer (2022) directly addressed Chinese legal texts' structural peculiarities – proving that scaled attention mechanisms could overcome jurisdiction-specific document length challenges unmanageable by earlier RNN models \cite{xiao2021lawformer}.

\textbf{Paradigm Shift(2023-2025)} studies collectively signify the critical transition:  Cui et al.'s ChatLaw (2023) MoE architecture exemplified domain-optimized scaling – where specialized legal expert modules dynamically route tasks within trillion-parameter systems \cite{cui2023chatlaw, cui2024chatlaw}; Wang et al.'s LLM-GNN fusion (2024) enabled multi-modal legal judgment prediction by combining textual reasoning with case relationship graphs \cite{wang2024causality}.

The representative works outlined above trace the developmental trajectory of legal AI in the LLM era. A more comprehensive and systematic literature review, organized through the Toulmin model-based classification framework, will be presented in Section 3.

\section{LLM-enhanced Legal Reasoning}

The core of judicial work is the justification of judicial decisions \cite{hohfeld1917fundamental}. The essence of this process lies in legal reasoning. In this case, Section 3 utilizes the legal reasoning model to elucidate the advancements in the legal LLMs field for systematicness, effectiveness and comprehensiveness. The application of this model highlights the latest theoretical interpretation of the formalization of legal reasoning within the context of the trend from symbolism to connectionism. Legal reasoning can be categorized into formal reasoning and substantive reasoning \cite{levi2013introduction}. Compared with formal reasoning, substantive reasoning models are more suitable for refining various legal tasks in the legal process because of their effectiveness in confirming major and minor premises, achieving external justification. Therefore, Section 3 adopts the widely accepted Toulmin model in substantive reasoning to model the LLM-enhanced legal reasoning process and discusses in detail the corresponding legal tasks in each part of the Toulmin model and the research progress of their operation under the enhancement of LLMs. %First, Section 3.1 introduces the method of decomposing large-scale model legal tasks on the basis of the Toulmin model, a substantive legal reasoning model. On this basis, the main content of the following sections in Section 3 is further introduced at the end of Section 3.1.

\subsection{LLM-assisted Legal Reasoning Structure}

Section 3.1 introduces the method of decomposing large-scale model legal tasks on the basis of the Toulmin model, a substantive legal reasoning model. To complete a comprehensive legal reasoning process, both formal and informal logic play irreplaceable roles. Formal reasoning, which in the legal field is generally judicial syllogism, ensures the completion of internal justification \cite{wroblewski1974legal}. Moreover, informal logic, which is commonly represented by the Toulmin model in the legal domain, ensures the completion of external justification. We first introduce the Toulmin model and elucidate the relationship between judicial syllogism and the Toulmin model.

\begin{figure}
  \centering
  \includegraphics[width=0.25\textwidth]{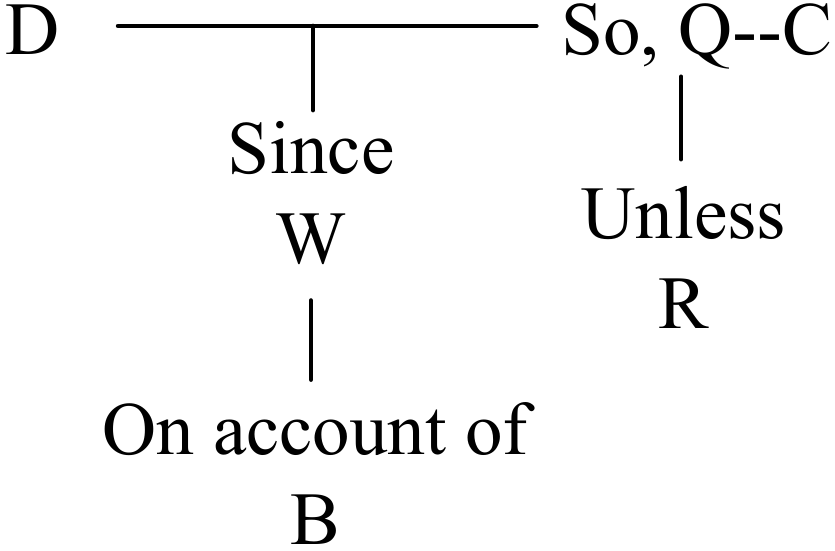}
  \caption{Framework of Toulmin Model.}\label{Fig_Toulmin}
\end{figure}

%To illustrate that LLMs play a role mainly in informal reasoning and that the Toulmin model can effectively summarize the application and development of LLMs in the legal field, it is necessary to elucidate the relationship between judicial syllogism and the Toulmin model. In the legal reasoning field, formal reasoning serves as the vehicle for conveying the content of substantive reasoning. This interaction addresses two of the most critical challenges in the reasoning process: ignorance of facts and ambiguity of objectives \cite{hart2012concept}. Therefore, how the Toulmin model in substantive reasoning fills in judicial syllogism in formal reasoning largely determines the correctness of legal decisions \cite{maccormick1994legal}.

As shown in Fig. \ref{Fig_Toulmin}, the Toulmin model dissects arguments into six key components—Data (D), Warrants (W), Backing (B), Qualifiers (Q), Rebuttals (R), and Claims (C)—and the entire argument process is a dynamic, adjustable structure that allows for adjustments and optimizations when faced with new evidence or arguments \cite{toulmin2003uses}. In the Toulmin model, the claim is the core of the argument, representing the main point that needs to be supported. To make the claim persuasive, it must rely on data, that is, specific facts, evidence, or examples that provide the foundation and support for the claim. The connection between the data and the claim is established through the warrant, which explains the logical relationship between the data and the claim, clarifying why the data can support the claim \cite{kneupper1978teaching}. Moreover, the validity of the data often requires additional supporting material, which is the Backing. Backing provides further evidence or authoritative opinions for the warrant, enhancing the credibility of the argument. To avoid overly absolute statements, Qualifiers are introduced to specify the scope and strength of the claim, preventing overgeneralization or misinterpretation of the argument \cite{hitchcock2005good}. Finally, the rebuttal is used to address potential counterarguments or exceptions. It acknowledges the challenges that the argument may face and provides corresponding explanations or evidence to weaken these challenges.

\begin{figure}
  \centering
  \includegraphics[width=\textwidth]{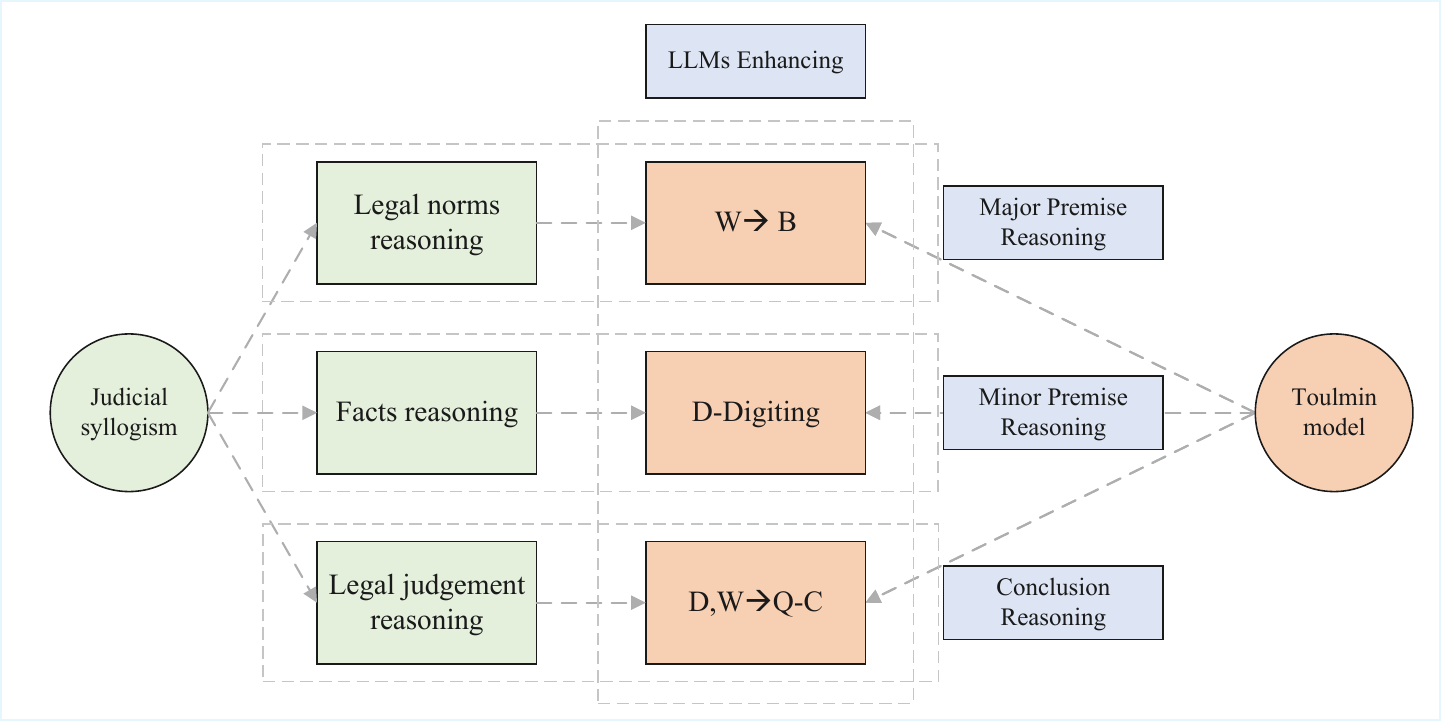}
  \caption{Relationships between formal reasoning and informal reasoning.}\label{Fig_Relationship}
\end{figure}

In judicial syllogism, legal rules are often the major premises that need to be discovered or determined. In such cases, the process of obtaining legal rules is a process of legal reasoning. At the same time, case facts are minor premises that involve a process of discovery or reasoning. Therefore, strictly speaking, legal reasoning encompasses three aspects: the reasoning of legal norms, the reasoning of facts, and the reasoning of judicial decisions \cite{dworkin1986law}, which can correspond to parts of the Toulmin model. As shown in Fig. \ref{Fig_Relationship}, LLMs facilitate the three crucial aspects of judicial syllogism by assisting with various tasks in the informal reasoning of the Toulmin model, thereby achieving both external and internal justification and completing a comprehensive legal reasoning process. Therefore, this section focus on which tasks within each component of the Toulmin model can be accomplished by LLMs and the progress of relevant research to assess the extent to which LLMs can assist in legal reasoning. %The large model-assisted judicial work of Section I can be organically integrated according to the Toulmin model, thereby coordinating the entire field of legal large language model research: 

\begin{figure}
  \centering
  \includegraphics[width=\textwidth]{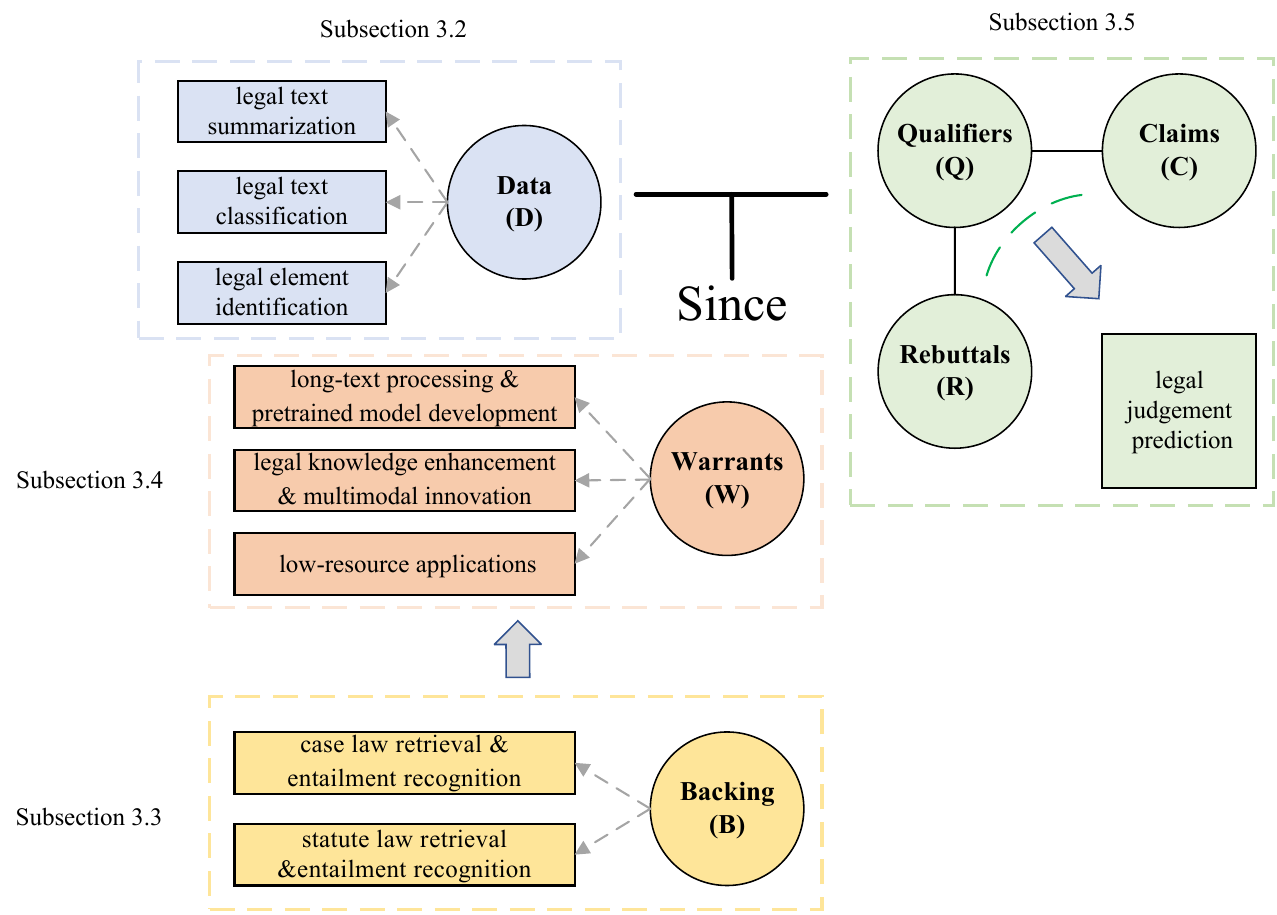}
  \caption{Decomposition of Legal Reasoning Tasks Based on LLMs.}\label{Fig_Decomposition}
\end{figure}

As indicated in Fig. \ref{Fig_Decomposition}, for each of these components, LLMs are capable of performing relevant tasks to assist legal practitioners in conducting legal reasoning. The integration of LLMs into the legal domain has significantly advanced the capabilities of legal text processing, analysis, and prediction, offering robust ability for enhancing legal reasoning in the Toulmin framework. This section provides a detailed analysis of the roles and contributions of LLMs in each component, emphasizing their technical and strategic significance in the legal context.

Based on Fig. \ref{Fig_Decomposition}, we systematically introduce the following tasks. In Section 3.2, the focus is on data processing, where LLMs efficiently handle large-scale legal texts through summarization, classification, and element identification, exemplified by models that condense complex documents into actionable insights. In Section 3.3, legal backing is addressed through case law and statute retrieval, supported by benchmarks to ensure that claims are grounded in authoritative precedents and statutory provisions. Section 3.4 emphasizes warrant optimization, leveraging long-text analysis, pretrained model fine-tuning, and multimodal integration to uncover implicit legal principles and enhance low-resource adaptability. Finally, Section 3.5 tackles argument formulation, enabling LLMs to generate and evaluate claims and qualifiers, and rebuttals while analyzing logical coherence to refine legal reasoning.

\subsection{LLM-assisted Legal Data Processing}

In this subsection, legal data processing tasks can be divided into three primary categories: legal text summarization, legal text classification, and legal element identification. %Before the advent of LLMs, legal text summarization was often achieved through manual abstraction or simple text extraction methods, which are time-consuming and prone to errors. Legal text classification relies mainly on manual tagging or traditional machine learning algorithms, which require a large amount of labeled data and have limited classification accuracy. Legal element identification is usually completed by lawyers manually reviewing documents, which is inefficient and inconsistent \cite{palau2009argumentation, bent2023large, siino2025exploring}. These traditional methods face several challenges, such as high labor costs, low efficiency, and limited accuracy. LLMs have significantly improved these tasks. For legal text summarization, LLMs can generate concise and accurate summaries of lengthy legal documents through advanced natural language processing capabilities, greatly enhancing efficiency and comprehension for legal practitioners. For legal text classification, LLMs can automatically categorize legal documents into relevant areas of law on the basis of their semantic understanding, improving the organization and retrieval of legal documents and streamlining workflows within law firms. For legal element identification, LLMs can accurately pinpoint key clauses, obligations, and rights in contracts and other legal documents, ensuring accuracy and mitigating risks. The assistance of LLMs not only boosts productivity but also enhances the precision and consistency of legal data interpretation. For legal text summarization, there are extractive summarization methods that select key sentences from the original text and abstractive summarization methods that generate new sentences to summarize the main content. For legal text classification, methods based on fine-tuning pretrained LLMs and methods that use LLMs to generate features for traditional machine learning algorithms exist. For legal element identification, methods based on named entity recognition and methods that use LLMs to generate structured information from unstructured text exist \cite{siino2025exploring}.

\subsubsection{Legal text summarization task}

Legal text summarization aims to extract key information from lengthy legal documents through automated technology, generating concise and accurate summaries to assist legal professionals in efficiently handling a large volume of legal texts. Legal text summarization research initially employed neural network methods such as word embeddings \cite{mikolov2013efficient, pennington2014glove, devlin2019bert} and sentence embeddings \cite{kiros2015skip, hill2016learning, logeswaran2018sentence} for semantic capture, laying foundations exemplified by work on Indian judgments \cite{anand2022effective}. However, these methods often struggled with context and generalizability. Reinforcement learning (RL) subsequently offered improvements, with advances including novel reward functions for deep modeling training \cite{nguyen2021robust}. The field has been significantly transformed by LLMs and related innovations: projects now leverage LLMs like GPT-4 via prompt engineering or fine-tuning methods such as LoRa, with memory-efficient training methods like
gradient accumulation, gradient checkpointing \cite{liu2024low}, to directly generate summaries and extract information, especially when facing low-source situations \cite{pont2023legal, benedetto2025leveraging, deroy2024applicability}, while novel architectures like the "extract-then-assign" (ETA) method address challenges of length and data scarcity through improved dataset construction \cite{jain2024summarization}. To improve the evaluation, particularly for lengthy legal opinions, new dimensions such as argumentation coverage ,event description accuracy, date accuracy, event capture completeness, and language quality have been introduced \cite{elaraby2024adding, mao2024comparative}.

\subsubsection{Legal text classification task}

The legal text classification task aims to classify legal documents efficiently and accurately through automated technology, thereby assisting legal professionals in quickly locating and processing relevant cases \cite{boella2011using}. Although early legal text classification used primarily traditional machine learning, NLP models, and deep learning techniques - including random forests \cite{chen2022comparative} and argument mining methods \cite{palau2009argumentation} - to process documents, the landscape changed significantly with the adoption of advanced transformer architectures and LLMs. Building on foundational transformer applications to legal issues \cite{nguyen2022transformer}, domain-specific pre-trained models like RoBERTa and BigBird were leveraged to classify cases from the European Court of Human Rights \cite{imran2023classifying}, and fine-tuned GPT-3 demonstrated effectiveness in legal rule classification \cite{liga2023fine}. LLM capabilities were further enhanced through task dependency, contrastive learning for case outcome prediction \cite{tyss2023leveraging}, and zero-shot transfer learning for generalizability to unseen legal articles \cite{santosh2023zero}. Recent innovations include employing label models to integrate ChatGPT outputs and exploring prompt engineering for legal text entailment \cite{nguyen2024employing, bilgin2024exploring}, alongside hierarchical frameworks for classifying large unstructured legal documents \cite{prasad2024exploring}, collectively demonstrating substantial progress in addressing legal text complexity.

\subsubsection{Legal element identification task}

The legal element identification task aims to extract key elements from legal documents, which is crucial for various legal applications, such as case analysis and legal information retrieval.

The development of techniques for this task has evolved from traditional machine learning models to more advanced deep learning architectures. Initially, Smuadu et al. explored the NER problem in Romanian and German legal domains via datasets such as RONEC and LegalNERo and employed models such as BERT, BiLSTM, and CRF \cite{smuadu2022legal}. This work laid the foundation for further research on legal element identification by demonstrating the effectiveness of domain-specific models. Yin et al. subsequently proposed a conversational mixture of expert large language models that was designed with specific prompts for case characteristics, utilized datasets such as CAIL2019 and a provincial theft case dataset, and employed tools such as ChatGLM3-6B and DoRA \cite{KXTS202412012}. This study advances the field by introducing a more sophisticated and interactive approach to legal case element recognition. Furthermore, recent advancements have seen the development of approaches that directly design prompts and retrieval-augmented generation (RAG) systems on the basis of LLMs for legal element identification. These methods have demonstrated superior performance, as evidenced by their successful application in relevant tasks \cite{reji2024enhancing}. Such innovations underscore the growing potential of LLMs in enhancing the precision and efficiency of legal element extraction, marking a pivotal step forward in the integration of artificial intelligence within the legal domain.

Research in this area has clearly progressed from basic models to more complex and context-aware systems, with each study building upon the previous ones to improve the accuracy, efficiency of legal element identification and multi-domain legal text processing technologies \cite{moro2024multi}.

\subsection{LLM-assisted Legal Backing Digiting}

The backing digiting task aims to match the applicable backings for case adjudications. In common law jurisdictions, backings are case precedents, whereas in civil law jurisdictions, backings are statutory provisions. This task is highly dependent on information retrieval and textual entailment technologies for legal texts. Since 2011, the COLIEE competition has focused on legal text information retrieval and recognition of legal provisions and case datasets \cite{goebel2024overview}. Therefore, relevant research generally revolves around COLIEE competition tasks and datasets. The COLIEE competition usually consists of four tasks, which are divided into two major categories: case law and statute law, as shown in Table \ref{tab:coliee_tasks}.

\begin{table}[t]
    \centering
    \caption{COLIEE Competition Task.}
    \label{tab:coliee_tasks}
    \begin{tabular}{|p{2.5cm}|p{3.5cm}|p{6.5cm}|}
        \hline
        Category & Task & Description \\
        \hline
        \multirow{2}{*}{Case Law Tasks} 
            & Task 1: Case Law Information Retrieval 
            & Participants are required to retrieve relevant cases from a given case repository based on query cases. The challenge lies in identifying potential associations between cases rather than relying on simple pattern matching. \\
        \cline{2-3}
            & Task 2: Case Law Entailment Recognition 
            & Participants need to determine whether there is an entailment relationship between existing cases and unseen cases. \\
        \hline
        \multirow{2}{*}{Statute Law Tasks} 
            & Task 3: Statute Law Information Retrieval 
            & Participants are required to retrieve relevant legal provisions from statutes based on queries. \\
        \cline{2-3}
            & Task 4: Statute Law Entailment Judgment 
            & Participants need to determine whether a given legal statement is correct. \\
        \hline
    \end{tabular}
\end{table}
	
While traditional legal text retrieval and entailment relied on basic deep learning and information retrieval, struggling with semantic complexity and provision-case relationships, the advent of LLMs overcame these limitations through superior language understanding, leading current trends to focus on advancing LLM architectures/fine-tuning and strategically integrating them with traditional IR techniques for optimal domain-specific performance.

\subsubsection{Early exploration and foundation: Combining foundational deep learning techniques with traditional information retrieval models (2018--2020)}

Initial development of legal information retrieval technology explored traditional models and basic transformers, exemplified by Tran et al.'s encoded summarization approach, which hierarchically encoded documents into a vector space for phrase scoring, validated on COLIEE 2018 data \cite{tran2019building}. Building upon early exploration, researchers shifted focus to deep learning methods. Shao et al. demonstrated this potential in 2020 with the BERT-PLI model, leveraging BERT to capture semantic relationships at the paragraph level and aggregating these interactions to infer case relevance, achieving significant improvements on COLIEE 2019 \cite{shao2020bert}. Furthermore, hybrid approaches emerged, such as Rabelo et al.'s framework combining traditional similarity models for initial scoring with transformer models for entailment classification, highlighting the advantage of integrating both methodologies \cite{rabelo2019combining}.

\subsubsection{Development and expansion: Improving deep learning models and new frameworks (2021--2024)}

With the continuous development of LLMs and legal information retrieval technology, researchers have begun to explore more complex technologies and model architectures. Research in legal information retrieval has evolved from standalone deep learning models to increasingly sophisticated integrations with LLM. The initial approaches used custom-trained models such as BERT ensembles for statute/case law retrieval \cite{shao2021bert} and unsupervised frameworks such as CaseFormer for capturing case distinctions \cite{su2023caseformer}. This foundation catalyzed a transition toward fusing deep learning with enhanced retrieval paradigms.

Subsequent innovations integrated transformers with traditional techniques like BM25 for legal entailment tasks \cite{kim2022legal}, while attention mechanisms evolved into hierarchical architectures for long legal documents \cite{nguyen2024attentive}. This period saw prolific hybrid systems combining ordinal keyword IR with BERT-based retrieval \cite{yoshioka2022hukb}, and merging BERT with graph networks and custom encoders \cite{wehnert2022applying}. Gleichzeitig, a a critical evaluation emerged othe performance of the method \cite{hudzina2020information} and the efficacy of pre-trainingicacy of pre-training \cite{zheng2021does}, establishing benchmarks for complex model architectures.

\subsubsection{Innovation and future outlook}

Recent innovations in legal information retrieval span three key dimensions. In retrieval frameworks, researchers propose multi-stage approaches \cite{nguyen2025retrieve} integrating LLMs with traditional methods for statute and case law retrieval \cite{nguyen2024pushing}, alongside methods improving accuracy through fine-grained fact-provision matching \cite{ge2021learning} and optimized retrieval-relevance ranking pipelines \cite{li2024towards}. Addressing cross-domain optimization and user diversity, studies focus on reducing misinformation via knowledge-enhanced verification \cite{milanese2025fact}, mitigating positional bias in RAG systems \cite{cuconasu2025rag}, calibrating confidence for ambiguous queries \cite{shi2025ambiguity}, and leveraging user attention for relevance judgment \cite{shao2023understanding}. Model structure innovations introduce GNN-based representations (CaseGNN) for capturing legal structural semantics \cite{tang2024casegnn} and causal relationship learning frameworks for provision selection \cite{wang2024causality}. Concurrently, data augmentation and model fusion techniques, employing contrastive learning, zero-shot prompting \cite{bui2024data}, model ensembles \cite{vuong2024nowj, nguyen2024nowj}, entity-aware tuning \cite{onaga2024contribution}, and interpretable Chain-of-Thought prompting \cite{fujita2024llm}, significantly enhance performance.

Future research must prioritize enhancing robustness across diverse legal domains and user backgrounds \cite{shao2023understanding, milanese2025fact}, addressing challenges like cross-jurisdictional generalization and bias mitigation \cite{cuconasu2025rag}. Advancing model interpretability through techniques like CoT \cite{fujita2024llm} and structurally-aware representations \cite{tang2024casegnn, wang2024causality}, coupled with deeper integration of domain knowledge for semantic alignment \cite{ge2021learning}, will be crucial. Further exploration into sophisticated LLM fusion strategies \cite{bui2024data, vuong2024nowj, nguyen2024nowj} and reliable verification paradigms \cite{shi2025ambiguity, milanese2025fact} holds significant potential to improve the accuracy, reliability, and practical utility of legal information retrieval systems.

%\subsection{LLM-assisted Legal Warrant Reasoning Generation}

\subsection{LLM-assisted Legal Warrant Reasoning Generation}

Legal warrant reasoning generation is crucial for extracting and reasoning with backings that support credible legal arguments. As legal texts grow in complexity, the ability to generate warrants based on known backings becomes a key challenge. This subsection outlines the development of LLM-assisted warrant reasoning in three aspects: long-text processing and legal domain pretraining, legal knowledge and multimodal integration, and applications in low-resource settings with future outlooks.

\subsubsection{Long-text processing and pretrained model development in the legal domain}

Long document processing is a foundational challenge in legal NLP. Lawformer \cite{xiao2021lawformer}, based on Longformer, demonstrated effective handling of long Chinese legal texts and outperformed prior models in tasks like judgment prediction. Its release as part of LegalPLMs provided a strong pretraining baseline, though it lacked explicit legal knowledge integration. Later models focused on domain adaptation. Lawyer LLaMA \cite{huang2023lawyer} fine-tuned LLaMA for legal Q\&A, reducing hallucinations but limited by narrow training data. DISC-LawLLM \cite{yue2023disc} introduced syllogistic prompting and retrieval-augmented mechanisms, signaling a shift to task-oriented legal LLMs. Standardized benchmarks emerged to assess these models. LawBench \cite{fei2023lawbench} evaluated Chinese legal tasks using OpenCompass, while LEGALBENCH \cite{guha2024legalbench} covered 162 cross-jurisdictional reasoning tasks. These benchmarks revealed limitations of general LLMs and set evaluation standards for legal AI \cite{fei2023lawbench, guha2024legalbench}.

\subsubsection{Legal knowledge enhancement and multimodal method innovation}

In 2024, structured legal knowledge integration became key. ChatLaw \cite{cui2024chatlaw} employed a MoE framework and knowledge graphs, reducing hallucinations by $38\%$ and improving citation-based reasoning. CBR-RAG \cite{wiratunga2024cbr} combined precedent retrieval with generation, boosting cross-jurisdictional Q\&A accuracy by $17.3\%$. InternLM-Law \cite{fei2024internlm} achieved state-of-the-art LawBench performance through two-stage fine-tuning. Ghosh et al. \cite{ghosh2024human} combined LLaMA-2 with expert-curated graphs, achieving $91\%$ precision in Indian legal document tasks, underscoring the value of human-in-the-loop systems. Hallucination analysis matured with Large Legal Fictions \cite{dahl2024large}, identifying $42\%$ error rates across 12 jurisdictions, and enabling mitigation evaluation via the Harvard Dataverse. Gray et al. \cite{gray2024empirical} enhanced empirical legal research using GPT-4 for factor annotation, tripling efficiency with $98\%$ consistency.

\subsubsection{Low-resource scenario applications and future technical directions}

LLMs have shown promise in low-resource contexts. Maree et al. \cite{maree2024transforming} used ChatGPT and LlamaIndex for legal Q\&A in Palestinian cooperatives, achieving $83\%$ answer relevance. Deng et al. \cite{deng2023syllogistic} introduced the SLJA Dataset, validating ChatGLM's effectiveness on 11,239 annotated criminal cases. Future directions include: (1) multimodal integration of text, knowledge graphs, and evidence \cite{wiratunga2024cbr, wiratunga2024cbr} and improvement of explainability, and resistance to AI-generated content interference using new architecture \cite{zeng2025rosilc}; (2) legal-specific generative pretraining to address data sparsity \cite{cui2024chatlaw, xiao2021lawformer, huang2023lawyer, yue2023disc, zhou2024lawgpt}; and (3) evolving evaluation systems such as LegalBench and 's multilingual expansion \cite{guha2024legalbench}. These trends signal a shift from supportive tools to robust legal decision-support systems.

\subsection{LLM-assisted Legal Judgment Prediction with Qualifiers}

In this subsection, we present an overview of the evolution and current landscape of legal judgment prediction (LJP). LJP has evolved from early statistical methods such as SVMs and logistic regression \cite{strickson2020legal} to deep learning models utilizing multistage case representation \cite{ma2020judgment, ma2021legal, prasad2023irit_iris_c}, further enhanced by reinforcement learning \cite{lyu2022improving}, multitask learning \cite{xu2020multi, huang2021dependency, yang2022mve}, and knowledge-augmented techniques via event extraction \cite{feng2022legal}, contrastive learning \cite{liu2022augmenting, zhang2023contrastive}, and logic-rule injection \cite{gan2021judgment}. While these models improved prediction accuracy and interpretability, they often struggled with generalization across languages and legal systems. The introduction of LLMs marked a turning point in LJP research, enabling models to better understand long-form legal texts and reason over complex legal structures, as evidenced by benchmarks like LexGLUE \cite{chalkidis2021lexglue} and multilingual datasets such as those from the Swiss Federal Supreme Court \cite{niklaus2021swiss} and LBOX OPEN \cite{hwang2022multi}, which laid the groundwork for evaluating pretrained models in legal contexts.

In the initial investigations of LLM-aided LJP, researchers concentrated on benchmarking fundamental performance and intrinsic limitations in statutory prediction and outcome forecasting, exposing both potential and challenges in real-world judicial settings. Shaurya Vats et al. \cite{vats2023llms} applied LLMs to Indian Supreme Court cases, observing superior statutory prediction accuracy but inferior outcome prediction compared to baselines, while uncovering gender and religious biases in model outputs. Ruihao Shui et al. \cite{shui2023comprehensive} proposed a practical LLM-based baseline solution, systematically evaluating LLM capabilities and demonstrating that analogous cases and label candidates boost performance, though weaker LLMs failed to exploit advanced information-retrieval systems; collectively, these studies highlighted performance bottlenecks and social-bias risks, furnishing an empirical foundation for subsequent refinements.

To address these bottlenecks, scholars devised novel frameworks for enhanced accuracy and interpretability through structured reasoning and model fusion. Shubham Kumar Nigam et al. \cite{nigam2024rethinking} employed LLMs like GPT-3.5 Turbo in Indian judicial contexts, simulating realistic courtroom settings with case facts and precedents to reveal superior performance and introduce clarity and relevance metrics. Chenlong Deng et al. \cite{deng2024enabling} proposed the ADAPT framework, decomposing facts and distinguishing charges to improve accuracy on ambiguous cases. Xia Yangbin et al. \cite{xia2024legal} introduced the LLG-Judger framework combining LLMs with graph neural networks, Zhang Yue et al. \cite{zhang2025rljp} developed a framework enhanced by first-order logic rules and comparative learning, and Bin Wei et al. \cite{wei2025llms} proposed a neuro-symbolic framework integrating logical rules with deep learning for civil-case adjudication—collectively marking the evolution toward refined reasoning and robustness in intricate scenarios.

In advanced analysis, researchers optimized practical deployment issues like information retrieval and noise suppression to improve stability and consistency. Ruihao Shui et al. \cite{shui2023comprehensive} re-evaluated LLM performance, confirming that analogous cases and label candidates enhance accuracy but noting retrieval systems alone may outperform combined approaches; follow-up experiments showed label-candidate augmentation improves consistency, while noisy cases disproportionately degrade smaller models, underscoring the need to balance retrieval efficiency with noise mitigation. Recent efforts have directly integrated LLMs into LJP pipelines, such as Wang et al.'s LegalReasoner framework \cite{wang2024legalreasoner}, a multistage architecture leveraging structured legal knowledge to enhance accuracy, interpretability, and adaptability to jurisdictional variations. Beyond binary prediction, LLMs now aim for qualified judgment outcomes using frameworks like the Toulmin Model to generate contextually grounded conclusions, such as nuanced sentencing decisions.

Looking forward, LLM-assisted LJP will benefit from cross-disciplinary collaboration integrating AI, law, and social sciences, as highlighted in surveys by Cui et al. \cite{cui2023survey} and Medvedeva et al. \cite{medvedeva2023rethinking}, emphasizing needs for richer datasets, multilingual resources, and improved interpretability to evolve LLMs into robust reasoning agents. In 2025, novel methods have been approached \cite{shao2025law}, such as JuriSim \cite{he2025simulating} and ILJR \cite{li2025basis}, incorporating the knowledge of judicial trial logic and other multi-source knowledge to manage the LJP task.

\subsection{Toolbox and Datasets}

We summarize the legal LLM tools and datasets mentioned in Section 3 in Table \ref{tab:legal-toolbox} and Table \ref{tab:toolbox-dataset}:

\begin{table}[htbp]
\centering
\caption{Legal LLM Toolbox}
\label{tab:legal-toolbox}
\resizebox{\textwidth}{!}{
\begin{tabular}{llll}
\toprule
\textbf{Name} & \textbf{Type} & \textbf{Creator} & \textbf{Reference} \\ 
\hline
Law2Vec & Word Embedding & Chalkidis et al. & \cite{chalkidis2017deep} \\
Legal-BERT & Pretrained Model & Chalkidis et al. & \cite{chalkidis2019neural} \\
MVE-FLK & Multi-task Legal LJP & Yang et al. & \cite{yang2022mve} \\
Lawformer & Long-text Processor & Xiao et al. & \cite{xiao2021lawformer} \\
ChatLaw & MoE Architecture & Cui et al. & \cite{cui2024chatlaw} \\
LawGPT & Fine-Tuning Model & Zhou et al. & \cite{zhou2024lawgpt} \\
DISC-LawLLM & Reasoning Framework & Yue et al. & \cite{yue2023disc} \\
InternLM-Law & Domain-specific LLM & Fei et al. & \cite{fei2024internlm} \\
Lawyer LLaMA & Legal QA System & Huang et al. & \cite{huang2023lawyer} \\
CaseGNN & Graph-based Retrieval & Tang et al. & \cite{tang2024casegnn} \\
LegalReasoner & Judgment Predictor & Wang et al. & \cite{wang2024legalreasoner} \\
ADAPT & Fact-charge Decoupler & Deng et al. & \cite{deng2024enabling} \\
LLG-Judger & GNN-LLM Fusion & Yangbin et al. & \cite{yang2024large} \\
CBR-RAG & Precedent Retriever & Wiratunga et al. & \cite{wiratunga2024cbr} \\
ETA Method & Abstractive Summarization & Jain et al. & \cite{jain2024summarization} \\
BERT-PLI & Statute Retriever & Shao et al. & \cite{shao2021bert} \\
Caseformer & Unsupervised Pretraining & Su et al. & \cite{su2023caseformer} \\
MLMN & Provision Matcher & Ge et al. & \cite{ge2021learning} \\
ContractMind & Agreement Drafting & Zeng et al. & \cite{zeng2025contractmind} \\
Syllogistic Prompting & Reasoning Enhancer & Deng et al. & \cite{deng2023syllogistic} \\
ERNIE 3.0 & Knowledge-enhanced LLM & Sun et al. & \cite{sun2021ernie} \\
Longformer & Sparse Attention & Beltagy et al. & \cite{beltagy2020longformer} \\
ELECTRA & Efficient Pretraining & Clark et al. & \cite{clark2020electra} \\
CTM & Judgement Prediction Framework & Liu et al. & \cite{liu2022augmenting} \\
BM25+Longformer & Joint Learning Approach & Nguyen et al. & \cite{nguyen2024nowj} \\
Multi-task Learning & Legal Named Entity Recognition & Smadu et al. & \cite{smuadu2022legal} \\
Causality-inspired Method & Relational Reasoning & Wang et al. & \cite{wang2024causality} \\
JuriSim & Dual Residual Attention for LJP & He et al. & \cite{he2025simulating} \\
ILJR & Interpretable Legal Judgment Reasoning & Liu et al. & \cite{li2025basis} \\
RoSiLC-RS & Robust Similar Legal Case Recommender System & Zeng et al. & \cite{zeng2025rosilc} \\
\bottomrule
\end{tabular}
}
\end{table}

\begin{table}[htbp]
\centering
\caption{Legal LLM Datasets}
\label{tab:toolbox-dataset}
\resizebox{\textwidth}{!}{%  % 等比例缩放至文本宽度
\begin{tabular}{llll}
\toprule
\textbf{Name} & \textbf{Task} & \textbf{Creator} & \textbf{Reference} \\ 
\midrule
COLIEE & Legal Information Retrieval & Goebel et al. & \cite{goebel2024overview} \\
LawBench & Summarization, Judgment Prediction, QA & Fei et al. & \cite{fei2023lawbench} \\
LEGALBENCH & Legal Reasoning Tasks & Guha et al. & \cite{guha2024legalbench} \\
LexGLUE & Multi-task Legal NLP & Chalkidis et al. & \cite{chalkidis2021lexglue} \\
CUAD & Contract Review, Obligation Identification & Hendrycks et al. & \cite{hendrycks2021cuad} \\
CAIL & Legal Element Recognition, Judgment Prediction & Hua et al. & \cite{KXTS202412012, wang2024causality} \\
Swiss-Judgment-Prediction & Multilingual Judgment Prediction & Niklaus et al. & \cite{niklaus2021swiss} \\
LBOX OPEN & Korean Legal Language Understanding & Hwang et al. & \cite{hwang2022multi} \\
SLJA Dataset & Syllogistic Reasoning, Legal Analysis & Deng et al. & \cite{deng2023syllogistic} \\
RealToxicityPrompts & Toxicity Detection, Bias Evaluation & Gehman et al. & \cite{gehman2020realtoxicityprompts} \\
RONEC and LegalNERo & Legal Named Entity Recognition & Smadu et al. & \cite{smuadu2022legal} \\
UK Case Law Dataset & Multi-task Legal NLP & Izzidien et al. & \cite{collenette2023explainable} \\
LAiW & IR, Legal Reasoning, Legal Application & Dai et al. & \cite{dai2023laiw} \\
RONEC & Romanian Named Entity Corpus & Dumitrescu et al. & \cite{dumitrescu2019introducing} \\
LegalDocML & Naming Convention & Dimyadi et al. & \cite{dimyadi2017evaluating} \\
LEXam & Legal Reasoning & Fan et al. & \cite{fan2025lexam} \\
NLJP& LJP, Classification & Chalkidis et al. & \cite{chalkidis2019neural} \\
\bottomrule
\end{tabular}%
}
\end{table}

\section{LLMs' Integrations in Dispute Resolution Procedures}

Section 4 explores how LLMs enhance both litigation and nonlitigation legal procedures, which resolve disputes through judicial processes in courts (governed by strict norms) or flexible alternatives like negotiation and arbitration, respectively. These models assist professionals in approximating legal "standard answers," with data-driven capabilities that inspire and assist, surpassing traditional tools \cite{martin2024better}. This section discusses integrating LLMs into these workflows, positioning human professionals as critical intermediaries for connecting AI to legal practice through human‒computer interaction.

\subsection{Applying LLMs in Litigation Procedures}

\subsubsection{Applying LLMs in civil litigation procedures}

In civil litigation procedures, LLMs provide comprehensive and efficient assistance to judges, lawyers, parties, and other litigation participants  through the various components of the Toulmin model, significantly enhancing the quality and efficiency of legal argumentation.

For judges, whose responsibilities include presiding over trials, ensuring fairness, and making fact-based rulings, LLMs can rapidly distill case facts and key legal provisions through legal text summarization and legal element identification techniques at the D part of the Toulmin model \cite{toulmin2003uses}. This helps judges efficiently grasp the overall picture of the case. For example, LLMs can summarize trial records, evidence materials, and legal provisions, providing judges with a clear factual framework and legal basis. In the W part of the Toulmin model, LLMs can enhance legal knowledge to assist judges in analyzing the logical relationship between case facts and legal provisions, ensuring accurate application of the law in their rulings \cite{maccormick1994legal}. Additionally, in terms of B, LLMs can quickly retrieve relevant precedents and legal provisions to provide authoritative support for judges' rulings, thereby enhancing the persuasiveness of their decisions \cite{hart2012concept,  goebel2024overview}.

Lawyers provide legal advice, draft documents, and present arguments in court. LLMs enhance their argumentation by summarizing legal texts, identifying key elements, and extracting relevant provisions, streamlining case preparation and document drafting \cite{siino2025exploring}. In the argumentation stage, LLMs, with their legal knowledge enhancement and multimodal innovation technologies, help lawyers construct logically rigorous legal arguments, ensuring that their claims are closely linked to legal bases \cite{cui2024chatlaw}. Moreover, LLMs can provide rich precedential and legal support through case law and statutory retrieval techniques, enhancing the credibility of their arguments \cite{nguyen2024pushing}.

Parties in civil litigation have the right to present their cases and evidence. LLMs can help parties better understand and express their claims by summarizing and refining their statements, enabling clear articulation of case facts and legal demands. Additionally, LLMs can provide basic legal knowledge through legal education functions, enabling parties to better protect their rights during the litigation process \cite{ghosh2024human}.

In civil litigation, LLMs enhance efficiency and precision. They help judges understand case facts, identify legal provisions, and apply precedents by analyzing legal texts. LLMs ensure judgments are logically coherent and grounded in law. For lawyers, LLMs streamline case preparation, from drafting pleadings to constructing arguments. Lay litigants benefit from LLM-generated summaries to articulate claims clearly. LLMs also assist witnesses and experts in preparing coherent testimonies. By bridging the knowledge gap, LLMs promote transparency and fairness in civil disputes.

\subsubsection{Applying LLMs in criminal litigation procedures}

The criminal litigation process involves various participating roles, each fulfilling specific functions and responsibilities. In the criminal justice systems of countries worldwide, key participants typically include judges, prosecutors, juries, defendants, defense attorneys, and witnesses such as expert witnesses. Judges preside over trials and apply the law based on the facts of the case to ensure the fairness and legality of the proceedings. Prosecutors initiate public prosecutions for criminal cases on behalf of the state, present charges, and provide sufficient evidence to substantiate the alleged crimes. Defense attorneys represent defendants, offer legal counsel, draft legal documents, and advocate for their clients. Additionally, witnesses provide testimony or professional opinions to assist fact-finders in determining the truth of the matter at issue \cite{grimm2017challenges}. 

In criminal litigation, LLMs provide targeted support to prosecutors and defense attorneys by enhancing legal argumentation through the Toulmin framework. Prosecutors leverage LLMs to draft indictments by summarizing facts and identifying essential crime elements \cite{siino2025exploring}, construct robust legal arguments connecting claims to statutory bases \cite{cui2024chatlaw}, and bolster arguments through precedent retrieval for evidential support \cite{nguyen2024pushing}. Defense attorneys utilize LLMs to rapidly identify case disputes and core defense strategies \cite{gray2024empirical}, critically analyze prosecution evidence to expose weaknesses in collection or authenticity \cite{wiratunga2024cbr}, discover exculpatory evidence through database mining \cite{maree2024transforming}, and generate targeted defense arguments grounded in facts and statutes \cite{deng2023syllogistic}. Additionally, LLMs assist witnesses in organizing testimony \cite{tang2024casegnn} and experts in rapidly accessing domain knowledge \cite{fujita2024llm}.

By applying the Toulmin model across these functions—structuring claims (C), grounding (D), warrants (W), backing (B), and rebuttals (R)—LLMs enhance argumentative rigor, factual relevance, and legal sufficiency. This deep integration ultimately improves procedural fairness and advances the pursuit of legally sound outcomes in criminal justice \cite{dworkin1986law}, while equipping all participants with more effective tools for truth-seeking.

\subsubsection{Applying LLMs in administrative litigation procedures}

In administrative litigation—distinguished by its focus on overseeing government power and placing the burden of proof on agencies—LLMs provide specialized support aligned with its unique objectives. For judges, they streamline review by summarizing administrative decisions and enforcement records while pinpointing critical legal elements like procedural compliance and evidentiary sufficiency \cite{elaraby2024adding}, and enhance rulings through precedent retrieval that analyzes judicial reasoning in analogous regulatory contexts \cite{prasad2024exploring}.

LLMs uniquely empower litigants: Plaintiffs gain clarity by distilling agency decisions and predicting outcomes through case law analysis \cite{nguyen2024pushing}, while government defendants leverage LLMs to meet strict proof burdens by rapidly generating compliance arguments (e.g., statutory adherence) and countering procedural challenges \cite{deng2023syllogistic}. Lawyers utilize LLMs to detect procedural flaws, clarify fact-law connections for conditional statutes (e.g., licensing rules), and strengthen arguments against administrative overreach \cite{siino2025exploring, reji2024enhancing}.

Crucially, LLMs generate conclusive, jurisdictionally tailored rulings (e.g., "Action unlawful due to Article X violation") by synthesizing factual, legal, and compliance criteria \cite{fei2023lawbench}. This capability uniquely advances administrative litigation’s core mission: balancing state authority with individual rights through rigorous legal-technical scrutiny absent in civil/criminal domains.

\subsection{Application of LLMs in Alternative Dispute Resolution Procedures}

In this section, we explore the integration of LLMs into Alternative Dispute Resolution (ADR) procedures across the three-stage nonlitigation framework (pre-conflict, conflict, dispute) \cite{nader1978disputing}. At the pre-conflict stage, LLMs enhance preventive services like contract review and legal advice by automating clause analysis, identifying risks, and providing proactive insights to prevent disputes. During the conflict stage, LLMs optimize negotiations through real-time risk assessments, automated drafting, and e-discovery for electronic evidence management. In the dispute resolution stage involving third parties (mediation or arbitration), LLMs assist mediators by generating structured agreements and support arbitrators through accelerated legal research, precedent analysis, enforceable award drafting, and ensuring cross-jurisdictional compliance. Collectively, LLMs streamline ADR workflows, improve decision-making, and promote efficient resolutions \cite{ogunde2024navigating}.

\subsubsection{Applying LLMs in the pre-conflict stage}

In legal practice, the pre-conflict stage is crucial. During this phase, preventive measures and proactive legal advice can reduce future disputes. Legal services at this stage include contract review, due diligence, and prelitigation consultation. LLMs enhance efficiency by automating clause analysis, identifying ambiguous clauses, and providing real-time feedback. In due diligence, LLMs extract critical information, automate compliance assessments, and forecast financial implications. In prelitigation consulting, LLMs offer comprehensive legal guidance to mitigate risks. 

Contract review is essential in legal practice, aiming to ensure clear, enforceable clauses aligned with best practices and minimizing disputes. Traditional methods rely on manual lawyer analysis, which is time-consuming, labor-intensive, and prone to errors. LLMs have transformed contract review by improving efficiency, enhancing clause analysis accuracy, increasing trust in AI recommendations, and offering real-time feedback for revisions. LLMs can quickly analyze contracts, identify ambiguous clauses, compliance issues, and cross-reference clauses with legal databases and standards 
\cite{zeng2025contractmind}.For example, The CUAD dataset provides high-quality labeled data for applying LLMs in contract review, enabling models to better understand and process contract texts  \cite{hendrycks2021cuad}.  Through dynamic interactive design, like the ContractMind system, LLMs can provide real-time AI recommendations, legal clauses, and case support, helping legal professionals efficiently review contracts by calibrating their trust in AI \cite{zeng2025contractmind}. LLMs can more accurately identify obligations, permissions, and prohibitions in contracts using CNNs and LSTM, providing legal professionals with comprehensive contract analysis \cite{graham2023natural}. These technological advancements have improved contract efficiency and accuracy, enhanced review quality and reliability through automation, and provided legal professionals with a more efficient solution \cite{mik2022much}.

During the due diligence process, ensuring transaction security and legality is the core objective. The application of LLMs is reshaping this traditional procedure by improving efficiency and accuracy. LLMs use methods like conditional random fields (CRFs) to process large volumes of unstructured data, such as financial reports and legal documents, enhancing information extraction \cite{roegiest2018dataset}. AI-driven frameworks employ machine learning algorithms, such as random forests, for financial data analysis to predict future company performance and assess financial health \cite{bedekar2024ai}. NLP automates legal compliance checks by reviewing documents and identifying risks, improving review efficiency and accuracy \cite{bedekar2024ai}. Additionally, LLMs analyze complex legal issues raised by administrative agencies and provide response suggestions, which are crucial for identifying legal risks \cite{deng2023syllogistic}. These technologies enhance due diligence efficiency and accuracy while providing decision-makers with deeper insights, ensuring secure and legal transactions.

In the pre-conflict stage, legal advisory services provide proactive guidance to prevent disputes. Due diligence involves evaluating a client's legal status, risks, and compliance with laws. This includes reviewing contracts and analyzing legal documents. Traditional consulting is often limited by lawyer availability and expertise, leading to inconsistent advice. LLMs offer consistent, high-quality legal insights by accessing extensive legal knowledge and precedents \cite{cui2023chatlaw}. LLMs analyze complex scenarios, interpret regulations, and generate tailored legal recommendations \cite{yue2023disc}, enhancing advisors' ability to deliver comprehensive advice \cite{fei2023lawbench}.
Moreover, LLMs significantly enhance real-time legal research and analysis capabilities, enabling advisors to promptly address client inquiries and emerging legal challenges \cite{fei2024internlm}. By integrating LLMs into legal advisory services, law firms can proactively identify and mitigate potential legal risks, minimizing disputes and fostering a more knowledgeable and resilient client base \cite{huang2023lawyer}. 

\subsubsection{Applying LLMs in the conflict stage}

In the conflict stage, the dispute can still be resolved through negotiation, a self-help remedy that does not involve third parties and is the least costly method. Negotiation often involves collecting and analyzing electronic evidence, so e-discovery is discussed in this subsection. As legal matters grow more complex and technology advances, LLMs are becoming key tools in the legal field. In e-discovery, LLMs enhance data processing efficiency and accuracy through technical improvements and hybrid methods while addressing ethical and privacy concerns. During negotiations, LLMs draft legally compliant agreements, interpret clauses, and assess legal risks in real-time, helping negotiators understand terms and create mutually beneficial agreements.

In the field of e-discovery, LLMs have emerged as a key solution to address data volume challenges and improve efficiency. Research has focused on three areas: technical optimization and ethical/privacy protection. For technical optimization, instruction tuning and techniques like LoRA/Q-LoRA enhance open-source model performance, surpassing commercial models in document relevance tasks and providing strong legal support for e-discovery \cite{pai2023exploration}. Ethical concerns are addressed through diverse training data and bias audits, while blockchain integration enhances data integrity and evidence chain management \cite{stjepanovic2024leveraging} to support clients' needs for non-discrimination and data accuracy in e-discovery processes. These advancements show that LLMs improve e-discovery efficiency, accuracy, and ethical handling of legal data.

In negotiations, negotiators must draft clear settlement agreements. LLMs can help create legally compliant drafts with precise language and accurate content \cite{huang2023lawyer}. Additionally, LLMs provide insights into relevant laws, assisting negotiators in assessing the legal risks and benefits of proposed terms \cite{fei2023lawbench}.

Negotiators must draft clear, legally sound agreements that reflect settlement terms. Well-drafted agreements reduce ambiguity, prevent disputes, and ensure enforceability. To do this effectively, negotiators need a strong grasp of legal terminology, contract structures, and relevant laws. LLMs can help by generating contract drafts aligned with legal standards and best practices. By analyzing extensive legal databases, LLMs ensure precise, unambiguous language tailored to the negotiation context. For instance, in agreements involving intellectual property, LLMs include appropriate clauses on ownership, licensing, and infringement protection \cite{cui2024chatlaw}. Additionally, LLMs identify inconsistencies and suggest improvements to enhance enforceability and minimize postnegotiation disputes.

Beyond the drafting process, negotiators must possess a comprehensive understanding of the legal implications associated with the terms under negotiation to make well-informed decisions and anticipate potential risks. Legal agreements often include complex clauses that can lead to unintended consequences if not thoroughly evaluated. In this context, LLMs function as an indispensable tool by delivering real-time legal insights and risk assessments. For example, when evaluating a limitation of liability clause, an LLM can rapidly analyze relevant legal precedents and statutory provisions to identify common pitfalls and enforcement challenges \cite{deroy2024applicability}. Additionally, LLMs benchmark proposed terms against industry standards, regulatory requirements, and judicial precedents, assisting negotiators in assessing the fairness and feasibility of the terms. By providing data-driven legal risk analysis, LLMs enable negotiators to construct agreements that are mutually advantageous and legally resilient, ensuring all parties clearly understand their respective rights and obligations before finalizing the deal.

\subsubsection{Applying LLMs in the dispute stage}

The application of LLMs in dispute resolution can enhance traditional legal processes by improving efficiency, accuracy, and fairness. As disputes grow more complex and diverse, LLMs provide advantages in arbitration and mediation by automating tasks, offering legal insights, and ensuring outcomes align with standards and interests.

(1)	Applying LLMs in arbitration: In arbitration, LLMs significantly support arbitrators by drafting enforceable awards with precise, legally compliant language to enhance enforceability \cite{gan2021judgment}. They expedite legal research by rapidly retrieving and analyzing relevant laws, regulations, and precedents—crucial for complex cases and aligned with trends showing 73\% of legal professionals planning AI adoption for such tasks \cite{kluwer20212021, xiao2021lawformer}. Furthermore, LLMs analyze case details against legal benchmarks, identify jurisdictional risks in international disputes, and provide actionable mitigation insights \cite{siino2025exploring}, thereby improving decision equity, process efficiency, and overall fairness in arbitration outcomes.

(2)	Applying LLMs in mediation: Mediation is a process that heavily depends on effective communication, sound legal reasoning, and the creation of structured settlement agreements. LLMs can substantially enhance the efficiency and fairness of mediation by supporting mediators in drafting clear, legally robust mediation agreements. These agreements must precisely reflect the interests and commitments of both parties while ensuring enforceability under applicable legal frameworks. LLMs are capable of generating well-structured draft agreements that adhere to established legal principles, minimize ambiguity, and mitigate the risk of future disputes. Additionally, they can analyze previous mediation cases with similar contexts to recommend widely accepted clauses, thereby assisting mediators in crafting balanced and comprehensive agreements that satisfy all stakeholders involved \cite{goswami2025incorporating}.

Another pivotal aspect of mediation is facilitating effective and constructive communication between the disputing parties. Mediators ensure both sides understand their legal positions, risks, and options. LLMs assist by summarizing pertinent laws, analyzing case law, and explaining complex legal terms impartially. This enables mediators to provide precise, objective information, promoting informed decision-making. Additionally, LLMs help structure sessions by creating agendas, encapsulating key points, and proposing compromise solutions based on data and patterns. By offering real-time insights and structured support, LLMs enhance mediators' capacity to guide parties toward resolutions \cite{ghosh2024human}.

LLMs can enhance post-mediation processes by automating the review of settlement agreements. They ensure alignment with laws and stakeholder interests by cross-checking terms against regulations, identifying risks, and proposing refinements. In cross-border cases, LLMs conduct comparative legal analyses for compliance across jurisdictions. By integrating LLMs, mediators improve efficiency, fairness, and the durability of settlements. Additionally, LLMs provide real-time legal insights and plain-language summaries to facilitate informed discussions and build trust in the mediation process \cite{head2024assessing}.

\section{The Collaboration of Technological Ethics and Legal Ethics}

This section analyzes legal perspectives on LLMs, focusing on two main areas: application issues and regulatory frameworks. The first part discusses risks, safety, discrimination, toxicity, and hallucination in LLMs. The second part examines the regulation of large models, including paradigm shifts, adaptations in legal fields, and systemic risks. Additionally, the legal profession must balance ethical considerations with technological use to ensure proper integration, as illustrated in Figure 5, setting the stage for discussions on human-LLM collaboration.

\begin{figure}
    \centering
    \includegraphics[width=0.9\linewidth]{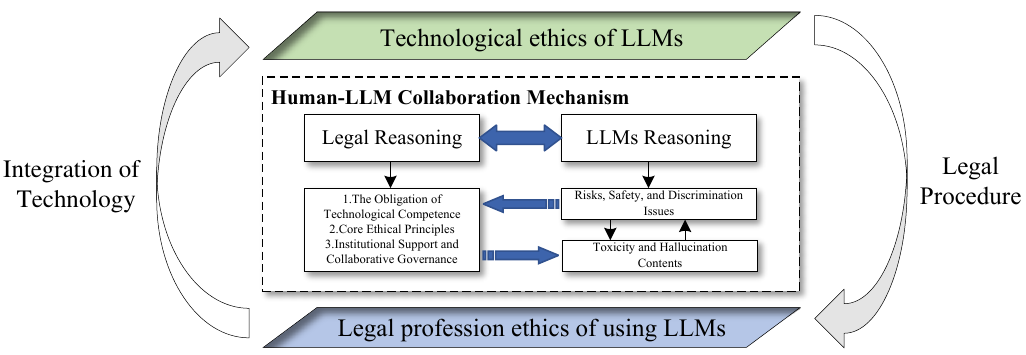}
    \caption{The Collaboration of Technological Ethics and Legal Ethics when applying LLMs in legal domain}
    \label{Fig_Collaboration}
\end{figure}

\subsection{Technological Ethics in the Application of LLMs}

Ethical and human alignment evaluations are crucial for legal AI. Guilherme F.C.F. Almeida et al.'s research found that the consistency between LLMs and human responses in moral and legal reasoning fluctuates, emphasizing that evaluation frameworks need to incorporate contextual factors to enhance credibility 
\cite{almeida2024exploring}. This direction highlights the ethical dimension of legal AI systems and provides key insights for developing more humanized models.

\subsubsection{Risks, Safety, and Discrimination Issues of LLMs}

In recent years, the rapid development and application of LLMs have prompted reflections on legal, ethical, and social risks. Blodgett et al. analyzed bias in NLP systems from theoretical and technical angles. Their 2020 study noted that quantitative research on bias often lacks clear motivation and normative reasoning, emphasizing the need to connect language analysis with social power structures to understand technology's harm to specific groups \cite{blodgett2020language}. In 2021, they critiqued fairness benchmark datasets for issues like ambiguous definitions, unverified assumptions, and inconsistent standards \cite{blodgett2021stereotyping}, proposing the integration of social science methods to redefine measurement frameworks. These insights provide a strong foundation for understanding technological bias.

Weidinger et al. introduced a six-dimensional classification system to expand ethical evaluation in risk frameworks. This system examines how language models may exacerbate social inequality through mechanisms like stereotype reinforcement, unequal resource distribution, and performance disparities. The authors also warned of derivative risks such as misinformation and privacy violations \cite{weidinger2021ethical}. Building on this, Bommasani et al. analyzed homogenization risks in foundational models like BERT and GPT-3. They emphasized that these models' multitask adaptability could propagate latent flaws into downstream applications, such as legal decision-making systems. Furthermore, the opacity of model outputs may lead to errors, especially in contexts where fairness verification is inadequate, potentially amplifying technical issues at an institutional level \cite{bommasani2021opportunities}.

In risk governance, scholars propose complementary solutions. The Bommasani team calls for interdisciplinary cooperation to break the "black box" of models and establish legal interpretability standards. Bender et al. focus on data sources, noting that unfiltered internet data in large models like GPT-3 internalizes biases. They recommend data cleaning and metadata recording to disrupt discriminatory content dissemination, crucial for sensitive applications like legal text generation \cite{bender2021dangers}. Both emphasize ethical integration throughout technology development.

These studies establish a logical framework from analyzing phenomena to governance strategies. Blodgett highlighted methodological flaws in bias research, Weidinger developed a framework for risk perception, and Bommasani and Bender proposed risk mitigation approaches from model architecture and data dimensions. Collectively, they emphasize aligning language model development with legal and ethical frameworks. Only through interdisciplinary collaboration, normative standards, and technological transparency can we balance technological innovation and social equity.

\subsubsection{Toxicity and Hallucination Contents of LLMs}

When examining the impact of large language models' toxic or hallucinatory outputs on the legal industry, we must address potential biases in processing specific group languages \cite{wu2025llm}. Blodgett et al. found that NLP tools are less accurate in identifying African American English (AAE) compared to standard English \cite{blodgett2016demographic, blodgett2017racial}. They suggested using mixed membership models to improve accuracy. These findings highlight NLP biases, which are critical for ensuring fairness and accuracy in legal texts. Additionally, Weidinger et al. identified six ethical and social risks of large language models, such as discrimination and misinformation, and proposed mitigation strategies emphasizing multi-party collaboration \cite{weidinger2021ethical}. These studies offer insights into reducing bias in NLP tools for the legal field.

Furthermore, the risks of language models spreading false information are critical concerns requiring legal attention. Zellers et al. introduced the Grover model, designed for generating and detecting neural fake news, emphasizing the importance of releasing robust generators publicly to enhance transparency and accountability \cite{zellers2019defending}. This research highlights the capabilities of language models in producing misleading content. Kreps et al. conducted three experiments examining the credibility of AI-generated text and its impact on public opinion, revealing that people often struggle to distinguish AI-generated from human-written news, with partisan bias affecting credibility assessments but minimally influencing policy perspectives \cite{kreps2022all}. Buchanan et al. explored GPT-3's potential for spreading false information, assessing its performance across disinformation tasks and demonstrating its ability to generate persuasive content with human assistance \cite{buchanan2021truth}. They outlined risks and proposed countermeasures for automated disinformation. Collectively, these findings warn the legal profession: the spread of false information could undermine judicial impartiality and erode public trust in the system.

Finally, detoxifying language models is a challenge the legal industry must consider. Gehman et al. developed the RealToxicityPrompts dataset to evaluate toxic content generation and analyze its sources in training data \cite{gehman2020realtoxicityprompts}. Welbl et al. found that while automatic detoxification methods reduce toxicity, they may overfilter marginalized groups and increase bias \cite{welbl2021challenges}. These studies offer insights for reducing toxicity in legal texts and ensuring fairness. Additionally, Gorwa et al. examined the technical and political challenges of algorithmic content moderation, including applications in copyright, terrorism, and toxic speech, emphasizing transparency, fairness, and accountability \cite{gorwa2020algorithmic}. 

These findings help the legal industry implement fair and unbiased moderation. LLMs can generate toxic or hallucinatory content, leading to issues such as discrimination against specific groups, spreading false information, and detoxification challenges. These studies offer valuable insights to ensure fairness and accuracy in legal texts while reducing bias.

\subsection{The Legal Profession Ethics of Legal Professions Using LLMs }

The rapid adoption of LLMs in the legal field has reshaped service delivery, efficiency, and posed challenges to Legal Professional Ethics (LPE). Traditional LPE, centered on obligations like competence, confidentiality, loyalty, diligence, and client communication, requires adaptation when addressing the risks and capabilities of LLMs. This section outlines the ethical guidelines for legal professionals using LLMs, focusing on establishing the "Obligation of Technological Competence" while ensuring other core ethical principles remain intact\footnote{The guidance principles issued by the Bar Council in January 2024 warned against blindly trusting the output of large language models and emphasized the risks of professional negligence, defamation, and data protection claims that may arise from improper use. In November 2023, the Bar Standards Board's guidance principles in the "Risk Outlook Report" pointed out that AI language models are prone to errors because they lack a concept of "reality," and may produce highly plausible but incorrect results. For specific cases, reference may be High Court of Justice, [The Avinide case], [2025] EWHC [1383] (QB).}. 

\subsubsection{The Obligation of Technological Competence: The Core Pillar of Lawyer Ethics in the LLM Era}

The "Obligation of Technological Competence" constitutes a core extension of the ethical duty of lawyers in the LLM era, evolving from the requirement of the American Bar Association (ABA) to understand the benefits and risks of technology \cite{baker2017beyond}. This obligation mandates that lawyers effectively utilize LLMs while maintaining critical oversight, requiring fundamental understanding of their principles, limitations, and task applicability \cite{mania2023legal, o2020smart}, coupled with proactive risk management of confidentiality breaches, inherent biases, and ethical risks \cite{zalewski2021basic, caserta2022sociology}.

Central to this duty is the rigorous supervision and verification of LLM outputs, as lawyers retain ultimate responsibility for all work product and must ensure accuracy, legal compliance, and protection of intellectual property \cite{o2020smart}. Additionally, the rapidly evolving landscape imposes a continuous learning imperative—through integrated ethics training and CLE—to maintain competence, justify fees, prevent malpractice, and counter de-professionalization pressures \cite{bialowolski2021does}.

\subsubsection{Upholding Core Ethical Principles: Confidentiality, Communication, Loyalty, and Diligence}

LLM-assisted legal practice introduces critical ethical obligations requiring strict adherence: (1) Mitigate confidentiality risks by evaluating third-party LLM providers' data policies, prioritizing secure deployment environments, inputting minimally redacted information, and obtaining client consent after risk disclosure \cite{dewan2005digital, van2017digital}; (2) Secure informed consent from clients regarding LLM usage scope, benefits, and confidentiality risks, ensuring transparency \cite{toren1975deprofessionalization}. (3) Uphold loyalty/diligence obligations by avoiding over-reliance that compromises personalized analysis or verification rigor; submitting unverified LLM outputs constitutes negligence \cite{abbott1981status}; (4) Prevent misrepresentation by meticulously verifying against primary sources to eliminate "hallucinated" precedents or plagiarized content, thus avoiding sanctions for misleading submissions \cite{caserta2022sociology}. These measures safeguard professional integrity amid technological adoption.

AI-generated legal arguments sometimes reference fictitious cases due to the system's failure to authenticate citations. Legal practitioners must verify AI-provided information against reputable sources. In some cases, attorneys are reminded of their responsibility to ensure the authenticity and originality of LLM-generated content \footnote{See R (Ayinde) v London Borough of Haringey; Al-Haroun v Qatar National Bank [2025] EWHC 1383 (Admin)}. 

\subsubsection{Institutional Support and Collaborative Governance: The Key Roles of Law Firms and Bar Associations}

Institutional support from law firms and bar associations is essential for ensuring ethical LLM implementation and upholding attorneys' technological competence obligations. Law firms must establish comprehensive governance frameworks—including approved-use policies, dedicated ethics oversight committees, and rigorous vendor/product assessments for security, compliance, and bias mitigation—while providing mandatory training on LLM risks, verification techniques, and ethical decision-making \cite{parra2024competent}. Additionally, firms must implement secure infrastructure, standardized client communication protocols (disclosures/consent), and external partnership safeguards to prevent unauthorized practice.

Bar associations play a critical standard-setting role by formally incorporating technological competence into professional conduct rules (e.g., ABA Model Rule 1.1) \cite{aba2024ai}, developing specialized CLE programs on legal tech ethics \cite{arrington2024vague}, and bridging the digital divide through equitable resource allocation. Proactive committee structures (e.g., ABA’s historical tech ethics initiatives \cite{kolb2016technology}, Shanghai's specialized committees \cite{shanghai2025committees}) enable authoritative guidance development, enforcement mechanisms for violations, and maintenance of centralized knowledge repositories to ensure uniform ethical standards and service quality across the profession.

\section{Roadmap and Future Directions}

Based on the comprehensive survey, future research directions for LLMs in law converge on three interconnected pillars:
\begin{itemize}
\item \textbf{Enhanced Legal Reasoning, Knowledge Integration \& Interpretability}: Future work requires multimodal fusion (text, images, charts) to deepen contextual analysis of legal evidence, alongside cross-jurisdictional adaptation through transfer learning for global deployment to reduce hallucinations and inconsistencies. Integrating structured legal knowledge via architectures and structure combining logical rules with neuro-symbolic computation \cite{wei2025llms} will reduce hallucinations while improving reasoning fidelity. Frameworks grounded in formal argumentation models (for example, Toulmin's warrant-backing structures) will drive accuracy insuch asks such as judgment prediction. Benchmarks such as JUREX-4E \cite{liu2025jurex}, LawBench \cite{fei2023lawbench} and LegalBench \cite{guha2024legalbench} will standardize the the evaluation, particularly for long-document processing and low-resource scenarios. Enhancing the interpretability of a model is also an important research direction. The development of more transparent and interpretable LLM models can help legal professionals better understand and trust the model's output. For example, visualizing the model's reasoning process or developing interpretive algorithms to extract key information can effectively increase the model's transparency and credibility. Optimizing long-document processing capabilities is also an important aspect of technical improvement. By improving methods for handling long documents, such as hierarchical encoding and clustering methods, or developing more efficient long-text segmentation techniques, the model's performance in processing complex legal documents can be significantly enhanced.

\item \textbf{Multi-Agent Workflow Augment}: By developing smarter legal question-answering systems, especially muti-agent workflow, more accurate and comprehensive legal consultations can be provided. For example, through multilingual support and legal knowledge enhancement, the system's performance in different languages and fields can be significantly improved, allowing it to better serve global users \cite{barron2025bridging}. Optimizing legal document processing is also an important practical direction. By optimizing tasks such as the generation, summarization, and classification of legal documents, the efficiency and quality of legal document processing can be significantly increased. For example, through multitask learning and contrastive learning, the model's performance in long-document classification and summarization tasks can be enhanced, thereby better meeting the needs of legal practice. Enhancing legal reasoning and judgment prediction is equally crucial. By combining legal knowledge and case data, the model's performance in legal reasoning and judgment prediction tasks can be significantly improved.

\item \textbf{Ethical \& Regulatory Co-Evolution}: Critical In the realm of legal ethics and social impact, reducing bias and discrimination is an important research direction. Through data cleaning, model optimization, and ethical review, bias and discrimination in the application of LLMs in the legal field can be effectively reduced. For example, through fairness analysis and bias detection technologies, the model's fairness and impartiality can be significantly enhanced, thereby better serving society. Enhancing transparency and interpretability is also an important ethical goal. By developing more transparent and interpretable models, legal professionals can better understand and trust the model's output. Establishing a legal ethics framework is also an important direction for development. Through interdisciplinary cooperation, ethical guidelines and regulatory frameworks can be formulated to effectively regulate the application of LLMs in the legal field, ensuring that technological development complies with legal and social ethical requirements. The argumentation framework of the legal profession based on the Toulmin model and its interaction with LLMs can help the legal profession focus more on the functions and goals of legal language \cite{shuy2002linguistic}, thereby fulfilling the professional ethical requirements for the technical competence of the legal profession.
\end{itemize}

\section{Conclusion}

The present work synthesizes the integration of LLMs into legal systems through an innovative dual-lens framework combining Toulmin argumentation structures and legal professional roles. It traces the field’s evolution from early symbolic AI and task-specific models to contemporary transformer-based LLMs, highlighting breakthroughs in contextual scalability , knowledge integration, and rigorous evaluation benchmarks, which are comprehensively shown in Table 2 and Table 3. The survey systematically categorizes advances across legal reasoning, dispute resolution workflows (litigation and alternative procedures), and ethical governance. Despite progress, critical challenges persist, including hallucination in legal claims, jurisdictional adaptation gaps in low-resource systems, explainability deficits in black-box reasoning, and ethical asymmetries in access and bias. Future work may prioritize multimodal evidence integration, dynamic rebuttal handling, and interdisciplinary frameworks aligning technical innovation with jurisprudential principles to realize robust, ethically grounded legal AI. To address the fundamental tension between algorithmic efficiency and judicial authority—a core controversy previously underexplored—we advocate for a legal-professional-centric strategy. This requires positioning LLMs as assistive tools rather than decision-makers, ensuring human oversight at critical junctures. Technical solutions must prioritize explainable reasoning paths. By centering legal expertise as the ultimate arbiter, we transform the "erosion" debate into a collaborative evolution: one where LLMs augment judicial wisdom while preserving the sacrosanct nature of legal authority.

\section*{Acknowledgment}

This research was supported by the Humanities and Social Sciences Research Project of the Ministry of Education of China (Grant No. 22YJC820054) and the National Social Science Fund of China (Grant No. 23BZX113).

\bibliographystyle{ACM-Reference-Format}
\bibliography{sample-base}

\end{document}